\def\year{2022}\relax
\documentclass[letterpaper]{article} 
\usepackage{aaai22}  
\usepackage{times}  
\usepackage{helvet}  
\usepackage{courier}  
\usepackage[hyphens]{url}  
\usepackage{graphicx} 
\usepackage{natbib}  
\usepackage{caption} 
\DeclareCaptionStyle{ruled}{labelfont=normalfont,labelsep=colon,strut=off} 
\usepackage{bm}
\usepackage{booktabs}
\usepackage{array}
\usepackage{multirow}
\usepackage{makecell}
\usepackage{arydshln}
\usepackage{graphicx}
\usepackage{color}
\usepackage{amsmath}
\usepackage{amssymb}
\usepackage{tikz}
\usepackage[noend]{algpseudocode}
\usepackage[linesnumbered, ruled, vlined]{algorithm2e}
\usepackage{pifont}
\newcolumntype{H}{>{\setbox0=\hbox\bgroup}c<{\egroup}@{}}

\newcommand{\triplet}[1]{$\langle$#1$\rangle$}
\newcommand{\ourDB}{KD-DTI}

\title{Discovering Drug-Target Interaction Knowledge from Biomedical Literature}
\author{
Yutai Hou$^1$, Yingce Xia$^2$, Lijun Wu$^2$, Shufang Xie$^2$, Yang Fan$^3$,  \\
        \textbf{Jinhua Zhu$^3$, Wanxiang Che$^1$, Tao Qin$^2$, Tie-Yan Liu$^2$} \\
        }
\affiliations{
    $^1$ Harbin Institute of Technology\quad$^2$ Microsoft Research \\
    $^3$ University of Science and Technology of China\\
    $^1$ \texttt{\{ythou,car\}@ir.hit.edu.cn},\quad$^3$ \texttt{\{fyabc,teslazhu\}@mail.ustc.edu.cn}\\
    $^2$ \texttt{\{yingce.xia,lijuwu,shufxi,taoqin,tyliu\}@microsoft.com} 

}

\begin{document}

\maketitle


\begin{abstract}
The \textbf{I}nteraction between \textbf{D}rugs and \textbf{T}argets (DTI) in human body plays a crucial role in biomedical science and applications. 
As millions of papers come out every year in the biomedical domain, automatically discovering DTI knowledge from biomedical literature, which are usually triplets about drugs, targets and their interaction,  becomes an urgent demand in the industry.
Existing methods of discovering biological knowledge are mainly extractive approaches that often require detailed annotations (e.g., all mentions of biological entities, relations between every two entity mentions, etc.).
However, it is difficult and costly to obtain sufficient annotations due to the requirement of expert knowledge from biomedical domains.  
To overcome these difficulties, we explore the first end-to-end solution for this task by using generative approaches. 
We regard the DTI triplets as a sequence and use a Transformer-based model to directly generate them without using the detailed annotations of entities and relations. 
Further, we propose a semi-supervised method, which leverages the aforementioned end-to-end model to filter unlabeled literature and label them. Experimental results show that our method significantly outperforms extractive baselines on DTI discovery.
We also create a dataset, \ourDB{}, to advance this task  and will release it to the community. 
\end{abstract}

\section{Introduction}
Discovering knowledge from biomedical literature is an important task for both research organizations and industrial companies. PubMED, one of the most famous search engines for biomedical literature,\footnote{https://pubmed.ncbi.nlm.nih.gov} has indexed more than $30M$ articles, and millions of new papers came out every year \cite{Landhuis2016}. It is impossible to manually check all the papers to obtain useful knowledge, which results in an urgent demand to automatically discover knowledge from the literature. 

The interaction between drugs and targets in the human body plays a crucial role in biomedical science and applications \cite{SACHDEV2019103159,wu2018network,yamanishi2008prediction}, e.g., drug discovery, drug repurposing, precision medicine, etc. In biomedical literature, a drug refers to any type of medication, ranging from small molecules like Aspirin, Penicillin to large molecules like Hepatitis B Vaccine. A target could be protein, enzyme or nucleic acid in our body, which binds the drugs we take. 
Drugs interact with targets in different ways. For example, Aspirin (drug) can inhibit (interaction) COX-1 (target), 
and Streptokinase (drug) can activate (interaction) Plasminogen (target). 
For simplicity, we call the triplet of Drug, Target and their Interaction as a ``DTI triplet''. 

\begin{figure*}[!t]
	\centering
	\begin{tikzpicture}
	\draw (0,0 ) node[inner sep=0] {\includegraphics[width=1.95\columnwidth, trim={0cm 7.45cm 1.2cm 0cm}, clip]{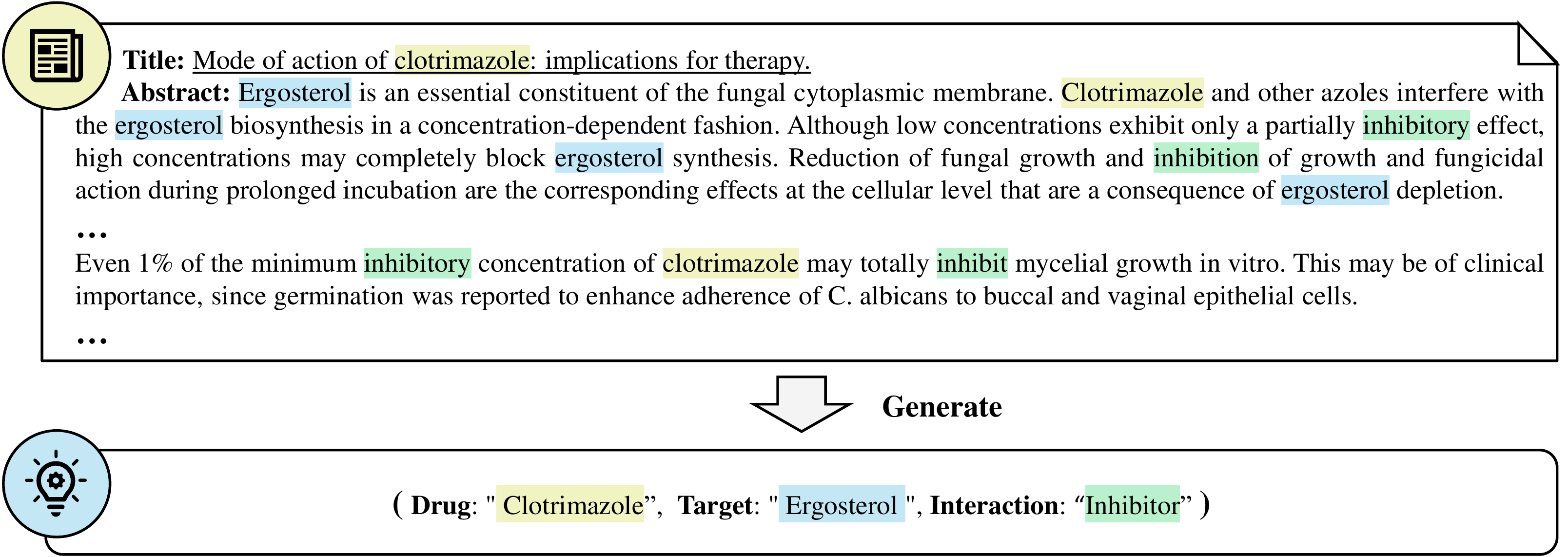}};
	\end{tikzpicture}
	\caption{\footnotesize
		Examples of drug-target-interaction knowledge discovery from literature.
	 }\label{fig:task_eg}
\end{figure*}

Discovering DTI triplets from biomedical papers is challenging. First, lots of terms and aliases (e.g., abbreviations, synonyms) exist in an article, but only a small set of them contributes to DTI triplets, which makes this task harder than conventional relation extraction from general text. 
As shown in Figure~\ref{fig:task_eg},  given the title and abstract of a paper, we want to discover the triplet \triplet{Clotrimazo, Ergostero, Inhibitor}. 
We can see that there are many terms like ``fungal cytoplasmic membrane’’, ``azoles'' and ``C. albicans’’,  which are not related to the triplet we want to discover and increase the difficulty of the task. 
Second, lots of DTI knowledge is expressed by multiple sentences~\cite{Verga2018SimultaneouslyST}, which is more difficult compared to the conventional relation extraction built upon one or two sentences~\cite{yao2019docred}. As shown in Figure~\ref{fig:task_eg}, no single sentence contains a complete DTI triplet.

Existing methods of discovering biological knowledge are mainly extractive approaches that usually sequentially use named entity recognition and entity relation extraction. 
Learning models for such pipeline requires detailed annotations, including all mentions of biological entities, relations between every two entity mentions and so on. 
However, it is difficult and costly to obtain sufficient annotations. 
On one hand, since the literature is usually long, the labeling workload is greatly increased compared to common knowledge extraction whose input is relatively short (e.g., a few sentences).
On the other, due to the diverse term-entity aliases and specialized expressions, identifying DTI triplets from papers requires expert knowledge of biomedical domains, as shown in Figure~\ref{fig:task_eg}. 

To rescue this task from the annotation difficulties, 
in this paper, we explore the first end-to-end solution.
We regard the DTI triplet as a sequence following the order of drug, interaction and target, and use a Transformer-based model for generation.
The document is fed into the encoder, and the DTI triplet is directly generated at the decoder.
Therefore, learning such model does not need to annotate every entity mentions and their relations.
The PubMedBERT~\cite{pubmedbert}, a model pre-trained on $30$M abstracts from biomedical literature, is applied to  the encoder and decoder of our model. 
Further, we use semi-supervised annotations to improve performance, where the end-to-end model trained on the limited labeled data is used to select/filter the literature and label them. 

Experimental results demonstrate that (1) generative models perform better than extractive ones and are more promising for this task; (2) leveraging unlabeled data can further boost the performance of generative models; (3) the performance of all the models is far from industrial demands,
even boosted by semi-supervised learning, 
which suggests that DTI discovery is a challenging task and calls for more research efforts from the machine learning and natural language processing community.

Finally, we provide a new dataset, \ourDB{}, for discovering DTI from documents. \ourDB{} is built upon DrugBank~\cite{Wishart2017} and Therapeutic Target Database (briefly, TTD) \cite{wang2020therapeutic}. As far as we know, \ourDB{} is the largest dataset for discovering DTI triplet from literature. The dataset contains  $12k$ training samples, $1k$ validation samples, and $1.1k$ test samples. We will release the data to the community.

Our  contributions are summarized as follows: 

(1) We explore the first end-to-end generative solution for extracting DTI triplet from biomedical literature, which shows the possibility of discovering DTI knowledge with much less annotation effort (\S \ref{sec:model_explore}). The method can be further enhanced by semi-supervised learning as we proposed (\S \ref{sec:semi})

(2) We create \ourDB{} (\S \ref{sec:ann}), the largest dataset for discovering Drug-Target-Interaction triplets from literature.
It contains $14$K labeled data and $3671$K words in total. 
We expect that such a dataset will advance the research of knowledge discovery from biomedical literature.

(3) We study the performance of current methods (\S \ref{sec:experiment}) and point out future directions for the DTI discovery task (\S \ref{sec:conclusion}).


\section{Our Method}\label{sec:model_explore}
In this section, we first introduce our end-to-end generative solution for DTI triplet discovery (\S \ref{sec:end2end}), and then describe how we utilize unlabeled data to further ease the difficulty of annotation (\S \ref{sec:semi}).

\subsection{End-to-end Triplet Generation}\label{sec:end2end}
To avoid labeling intermediate annotations (i.e., labels for entities mentions and relations between each pair of entities) and sequentially applying multiple models as extractive methods, we explore generative methods for this task. 
Specifically, we use a Transformer model \cite{Transformer}. The encoder of Transformer is used to encode the document (i.e., title and abstract), and the decoder of Transformer works for generating the DTI triplets. The output of the decoder follows the following format:
\begin{center}
$\texttt{<d>}$ drug$_1$ $\texttt{<i>}$ interaction$_1$ $\texttt{<t>}$ target$_1$ $\texttt{<d>}$ drug$_2$ $\texttt{<i>}$ interaction$_2$ $\texttt{<t>}$ target$_2$ $\cdots$,
\end{center}
where the drug, interaction and target are separated with special tokens $\texttt{<d>}$, $\texttt{<i>}$ and $\texttt{<t>}$, and all triplets are concatenated as a longer sequence. 

Recently, pre-trained language models, such as BERT \cite{devlin2018bert} for common domain and PubMedBERT~\cite{pubmedbert} for biomedical domain, achieve great success in NLP areas.
However, as suggested by previous work \cite{UNILM,DBLP:conf/iclr/ZhuXWHQZLL20}, 
directly using BERT to initialize parameters of generation models is not the best choice.\footnote{Using generative pre-trained models (e.g., GPT2 \cite{radford2019language}) may avoid such effects, and we leave it as future work.}
Therefore, we follow \cite{DBLP:conf/iclr/ZhuXWHQZLL20} to indirectly incorporate the pre-training models.

For encoder, we first use BERT to extract token representations  as $B = \{\bm{b}_1, \bm{b}_2, ..., \bm{b}_n\}$, where $\bm{b}_i$ is BERT embedding of the $i$-th token.
After that, given an $L$-layer Transformer encoder, we obtain final encoder output by incorporating additional attention over the features extracted by BERT. Mathematically, the outputs of the $l$-th encoder layer, $H_l$, is calculated as follows:
\begin{equation}
\begin{aligned}
	A_l & = \frac{1}{2} ({\rm Attn}(H_{l-1}, H_{l-1}) + {\rm Attn}(H_{l-1}, B)), \\
	\tilde{H}_l & =  {\rm LN}(H_{l-1} + A_l), \\
	H_l & = {\rm LN}(\tilde{H}_l + {\rm FFN}(\tilde{H}_l)), \\
\end{aligned}
\label{eq:encoder_side}
\end{equation}
where $H_0 = \{\bm{e}_1, \bm{e}_2, ..., \bm{e}_n\}$ is the embedding of all input tokens\footnote{$H_0$ in Eqn.\eqref{eq:encoder_side} and $H'_0$ in Eqn.\eqref{eq:decoder_side} are not the BERT embedding $B$, but the input embedding of the Transformer encoder/decoder, which are randomly initialized and learned during training.}.
${\rm LN}$ is the layer normalization. ${\rm Attn}(\bm{Q}, \bm{V})$ is the multi-head self-attention function operating on vector packages of queries $\bm{Q}$, values $\bm{V}$ (also used as keys). ${\rm FFN}$ is a position-wise feed-forward network.
We denote $R = H_L$ as the final representations of input tokens.


On the decoder side, we use the attention models over BERT features in a similar way to the encoder. The hidden state of $j$-th decoding step is calculated as:
\begin{equation}    
\begin{aligned}
    \tilde{A}'_l & = {\rm LN}(H'_{l-1} + {\rm Attn}(H'_{l-1}, H'_{l-1})), \\
	A'_l & = \frac{1}{2} ({\rm Attn}(\tilde{A}'_l, R) + {\rm Attn}(\tilde{A}'_l, B)), \\
	\tilde{H}'_l & =  {\rm LN}(\tilde{A}'_l + A'_l), \\
	H'_l & = {\rm LN}(\tilde{H}'_l + {\rm FFN}(\tilde{H}'_l)), 
\end{aligned}
\label{eq:decoder_side}
\end{equation}
where $H'_l$ is the output of the $l$-th decoding layer, which is a package of $j$ hidden state representations. $H'_0 = \{\bm{e}'_1, \bm{e}'_2, ..., \bm{e}'_j\}$ is the non-BERT embedding of all decoded tokens plus a \texttt{<SOS>} token (input of the first decoding step).

\subsection{Semi-supervised Learning}\label{sec:semi}
The end-to-end solution proposed in the previous section reduces the efforts of detailed annotation, like annotating where an entity starts/ends in a paper (even it does not contribute to the final DTI), or the evidence that how a relation is obtained. However, it is still cost to obtain a large amount of the (document, DTI triplet) pair, since such labeling requires solid biomedical background. 
To remedy this, we proposed a semi-supervised method based on our end-to-end model. The method consists of two steps, where we first filter the data based on rules, and then use the model in the previous section to refine the labels. We also discuss the relation with distance supervision.


\noindent(Step-1) {\em Rule-based filtration}: We download all available titles and abstracts from PubMED. For each document (i.e., title and abstract), we use ScispaCy \cite{neumann-etal-2019-scispacy}, an open-sourced NER tool to find out all possible drug and target entities, 
and use \texttt{FuzzyMatch} to search all mentioned interactions in the document\footnote{We collect commonly used interactions from DrugBank in advance, like inhibition, agonist, antagonist, etc.}.
Assume we extract $m$, $n$ and $p$ drugs, targets and interactions from document $A$. By enumerating all permutations, document $A$ corresponds to $m \times n \times p$ $\langle$drug, target, interaction$\rangle$ triplets. 
An underlying assumption is that the DTI triplets that appear more frequently across all documents have better confidence to be true.
We then count the numbers of occurrences of each DTI triplet and delete DTI triplets with less than $10$ occurrences.\footnote{According to our preliminary exploration, if we randomly select a drug, a target and an interaction from our dataset, most of those DTI triplets occur less than 4 times in all the PubMed papers.} We keep the documents that have at least one DTI triplet after the deletion. We eventually obtain a dataset with $139k$ documents, which is denoted as $\mathcal{D}_{\rm semi}$.

\noindent(Step-2) {\em Model-based labeling}: We propose a method based on knowledge distillation~\cite{hinton2015distilling} to refine the DTI triplet in $\mathcal{D}_{\rm semi}$. First, we use a generative model trained on supervised data to generate DTI triplets for each unlabeled document.
If the Transformer model does not generate any triplet for a document from $\mathcal{D}_{\rm semi}$, we remove the document from $\mathcal{D}_{\rm semi}$. Each remaining document in $\mathcal{D}_{\rm semi}$ is associated with at least one generated triplet and at least one pseudo triplet. If at least two elements (e.g., drug-target, drug-interaction, or target-interaction) of a pseudo triplet are the same as those of a generated triplet, we keep this document (and the matched triplet in $\mathcal{D}_{\rm semi}$); otherwise, we delete it. After filtration, there are $5.8k$ documents left in the dataset. Denote this dataset as $\mathcal{D}_{\rm KD}$. Note we will use the matched pseudo triplets in $\mathcal{D}_{\rm semi}$ for the following experiments; the generated triplets are only used for filtration, but not for model training. 

Another way to use unlabeled data is distance supervision~\cite{mintz2009distant} (DS). Given any DTI triplet $(d,t,i)$ from the labeled dataset, we use \texttt{FuzzyMatch} to search all $\bar{D}_j$ in $\mathcal{D}_{\rm semi}$. 
If we find mentions of both $d$ and $t$, we assign a pseudo label/triplet $(d,t,i)$ to $\bar{D}_{j}$. 
Denote the obtained dataset as $\mathcal{D}_{\rm DS}$, which has $15k$ samples. We empirically found that our proposed semi-supervised method is better than distance supervision, since DS does not take interactions into consideration.



\section{The Corpus}
In this section, we first introduce the acquisition of the proposed dataset \ourDB{} (\S \ref{sec:ann}), and then introduce its statistics and characteristics (\S \ref{sec:stats}). The dataset is available in the supplementary material and will be released later.


\subsection{Dataset Creation}\label{sec:ann}
\begin{table*}[t]
	\centering
	\footnotesize
	\renewcommand{\arraystretch}{1.3}
	\begin{tabular}{@{}  l c c c c c @{}}
	\toprule
	Dataset & \# Document & \# Relation & \# Sentence & \# Words & Knowledge \\
	\midrule
	CPI-DS \cite{doring2020automated} & N/A & 1 & 2,613 &  486k & Chemical Proteins Relation  \\
	BC5CDR \cite{li2016biocreative} & 1,500 & 1 & 11,089 &  282k & Chemical Disease Relation  \\
	ChemProt \cite{chemprot} & 2,432 & 5 & 24,923 & 650k & Chemical Proteins Relation \\
	\hdashline
	\ourDB{} & 14,256 & 66 & 139,810 & 3,671k & Drug Target Interaction \\
	\ourDB{} (semi) & 139,408 & 66 & 1,556,614 & 39,997k & Drug Target Interaction \\
	\bottomrule
\end{tabular}
\caption{
Statistic of datasets for discovering biomedical knowledge. ``\ourDB{} (semi)'' denotes semi-supervised dataset in \S \ref{sec:semi}. 
}
\label{tbl:basic_stat}
\end{table*}

\noindent{\bf Data collection}
The DTI triplets in our dataset come from two widely used databases, DrugBank \cite{Wishart2017} and Therapeutic Target Database (TTD) \cite{wang2020therapeutic}. (1) DrugBank is a pharmaceutical knowledge base that consists of proprietary authored content describing clinical-level information about drugs.\footnote{\url{https://go.drugbank.com/}} DrugBank covers $14,315$ drugs, $4,885$ targets, $63$ types of interactions and $18,866$ DTI triplets. (2) TTD\footnote{\url{http://db.idrblab.net/ttd/}} is a comprehensive collection of various types of drugs, which includes $37,316$ drugs, $3,419$ targets, $109$ interactions and $43,874$ DTI triplets. Given a DTI triplet, if the reference papers are provided and the abstracts of those papers are openly accessible, we record the triplet and the paper. As the first step, we only use the titles and abstracts of the reference papers. The dataset of this step is denoted as $\mathcal{D}$.

\noindent{\bf Data filtration}
As a starting point of structured DTI knowledge discovery, we are only interested in the document which contains enough information to discover a DTI triplet. However, in $\mathcal{D}$, 
some papers only generally describe some drugs and targets, in which the DTI triplets do not explicitly appear. 
Therefore, we heuristically filter out the samples in $\mathcal{D}$ by which we cannot obtain the associated DTI triplets. 
The basic idea is that we require that the drug, target, and interaction in a triplet should be all included in a paper (name or alias appear). 
After filtration, we obtain $1.3k$ documents as test set, $1k$ documents as the validation set, and $12k$ documents as the training set. 
The detailed filtration process is described in Appendix B.

\noindent{\bf Human verification} We then manually check all the samples in the test sets. We employ eleven annotators with Ph.D. background. Each (document, DTI triplets) pair is independently checked by two annotators. If their evaluation results are different, another two annotators are involved for discussions. We remove those difficult cases that a consensus is not reached after the discussions of four annotators. We eventually obtain $654$ (document, DTI triplets) pairs from DrugBank, and $505$ pairs from TTD for the test set.
We explicitly split them into two parts so that one could check the performance on different data sources.

\subsection{Comparisons with Previous Datasets}\label{sec:stats}

Table \ref{tbl:basic_stat} shows the statistics of our dataset as well as some related datasets. We have the following observations:

\noindent(1) In terms of data size (including numbers of documents, sentences and words), our dataset is much larger than previous datasets. 

\noindent(2) Although ChemProt and CPI-DS also focus on tasks in the biomedical domain, they do not directly serve for drug-target interaction discovery from literature. ChemProt mainly focuses on relation extraction and assumes that entities are given in advance, while our \ourDB{} is to discover DTI triplets (instead of relation only) from documents.  CPI-DS is to extract relation from single sentences, and thus is much easier than our task that tackles long documents.

\noindent(3) \ourDB{} includes a rich variety of relationship types that are not covered by previous datasets. ChemProt covers five relations, and CPI-DS and BC5CDR contain one relation only. In comparison, there are $66$ relations in our dataset.

\noindent(4) \ourDB{} is collected from more than one data source, i.e., DrugBank and TTD, which could be used to evaluate the generalization or transfer abilities of machine learning algorithms/models.

\section{Experiments}\label{sec:experiment}
In this section, we introduce several extractive baselines in \S \ref{sec:extractive_methods}, describe the  settings of  experiments in \S \ref{sec:exp_settings} and show the results of our end-to-end method and the results of using semi-supervised learning in \S \ref{sec:results_ours} and \S \ref{sec:results_semi}.


\subsection{Extractive Baselines}\label{sec:extractive_methods}
We compare with two extractive baselines: Cascade Relation extraction (CasRel), which is the state-of-the-art extractive method \cite{wei2020novel} and a pure NER method, which regards relation as a special entity.

\noindent{\em CasRel}: CasRel is a cascade tagging method that can jointly perform NER and RE. CasRel leverages BERT to extract representations for input sequences. To find out DTI triplets, CasRel first tags out all possible drugs (i.e., subject) of the input. After that, CasRel searches interactions (i.e, relation) and targets (i.e., objective) for the discovered drugs. For this purpose, we train a classifier for each relation, 
whose input is the discovered drug and the output is the position of the target, i.e., classifications of whether each token is the start or end token for the target phrase. 
The classifier is allowed to output null, indicating no target for this relation. 

To use CasRel, we obtain the named entity annotations of drugs and targets by searching the document with \texttt{FuzzyMatch}. Although CasRel achieved great success in standard relation extraction tasks like NYT \cite{riedel2010modeling} and WebNLG \cite{gardent2017creating}, in our setting, the annotations for all entities are automatically obtained without manually checking, which limits the performance of CasRel.





\noindent{\em Pure NER method}:
In biomedical literature, interactions often explicitly appear in documents with specific forms (e.g., noun, verb, past/present participle). Therefore, it is natural to regard the interaction as a special entity, and use a NER model to figure out the DTI triplets. For this purpose, after obtaining the mentions of drugs, targets and interactions using \texttt{FuzzyMatch}, we train a BERT-based NER model where interactions are also types of entity. During training, the BERT-based NER model is trained to 
predict the possibility of whether a token belongs to entity spans of drugs, targets and interactions.
At the inference phase, the trained model tags out token spans of drugs, targets and interactions.
For simplification, we choose the drug, target and interaction with the maximum probability 
to constitute the DTI triplet.
With this method, we can predict at most one DTI triplet for each document.

\subsection{Settings}\label{sec:exp_settings}

\begin{table*}[t]
\centering
\small
\renewcommand\arraystretch{1.2}
\begin{tabular}{l rrr rrr rrr}
\toprule
\multirow{2}{*}{\textbf{DrugBank}} & \multicolumn{3}{c}{\textbf{Triplet Level}} & \multicolumn{3}{c}{\textbf{Ontology Level}} & \multicolumn{3}{c}{\textbf{Entity Level (Acc.)}} \\
			\cmidrule(lr){2-4}
			\cmidrule(lr){5-7}
			\cmidrule(lr){8-10}
			 & \multicolumn{1}{c}{\textbf{F1} } & \multicolumn{1}{c}{\textbf{P} } & \multicolumn{1}{c}{\textbf{R} }  
			 & \multicolumn{1}{c}{\textbf{F1} } & \multicolumn{1}{c}{\textbf{P} } & \multicolumn{1}{c}{\textbf{R} }
			 &  \multicolumn{1}{c}{\textbf{Drug}} & \multicolumn{1}{c}{\textbf{Target}} & \multicolumn{1}{c}{\textbf{Interact}}  \\
			\midrule
			
            CasRel           & 15.42 & 13.74 & 17.57 & 18.62 & 20.74 & 16.89 & 27.12 & 23.14 & 31.32 \\
            Pure NER         & 18.20 & 19.11 & 17.37 &    17.25 & 19.40 & 15.54 &     60.72 & 35.25 & 79.26  \\
            Transformer          & 27.41 & 28.38 & 26.50 &    25.49 & 26.64 & 24.43 &     53.05 & 52.86 & 79.54 \\
            Transformer + BERT    & 30.32 & 31.46 & 29.26 & 29.50 & 31.64 & 27.64 & 53.00 & 55.27 & 79.47 \\
            Transformer + PubMedBERT    & 34.82 & 35.88 & 33.82 & 32.87 & 34.73 & 31.22 & 55.73 & 58.91 & 82.11 \\
            Transformer + BERT-attn    & 34.60 & 35.50 & 33.74 & 33.26 & 35.50 & 33.74 & 56.80 & 55.12 & 79.82 \\
            Transformer + PubMedBERT-attn    & 36.97 & 37.82 & 36.16 &    34.32 & 36.64 & 32.28 &     57.33 & 58.69 & 82.59 \\
\midrule
			\multirow{2}{*}{\textbf{TTD}} & \multicolumn{3}{c}{\textbf{Triplet Level}} & \multicolumn{3}{c}{\textbf{Ontology Level}} & \multicolumn{3}{c}{\textbf{Entity Level (Acc.)}} \\
			\cmidrule(lr){2-4}
			\cmidrule(lr){5-7}
			\cmidrule(lr){8-10}
			 & \multicolumn{1}{c}{\textbf{F1} } & \multicolumn{1}{c}{\textbf{P} } & \multicolumn{1}{c}{\textbf{R} }  
			 & \multicolumn{1}{c}{\textbf{F1} } & \multicolumn{1}{c}{\textbf{P} } & \multicolumn{1}{c}{\textbf{R} }
			 &  \multicolumn{1}{c}{\textbf{Drug}} & \multicolumn{1}{c}{\textbf{Target}} & \multicolumn{1}{c}{\textbf{Interact}}  \\
			\midrule
			
            CasRel           & 5.74 & 4.87 & 7.00 & 6.05 & 9.30 & 4.49 & 18.77 & 18.51 & 18.06 \\
            Pure NER        &  6.66 & 6.77 & 6.55 &    6.32 & 9.23 & 4.81 &     33.82 & 16.93 & 68.40  \\
            Transformer          & 6.32 & 6.73 & 5.96 &    5.75 & 6.41 & 5.22 &     10.89 & 56.53 & 87.43 \\
            Transformer + BERT    & 7.63 & 7.87 & 7.41 & 7.36 & 8.44 & 6.53 & 14.44 & 51.98 & 87.72  \\
            Transformer + PubMedBERT    & 7.81 & 8.28 & 7.41 & 7.11 & 7.83 & 6.52 & 12.83 & 58.47 & 86.93 \\
            Transformer + BERT-attn    & 8.34 & 8.42 & 8.27 & 7.59 & 8.14 & 7.10 & 15.46 & 53.37 & 87.03 \\
            Transformer + PubMedBERT-attn    & 8.88 & 9.21 & 8.57 &    7.87 & 8.83 & 7.10 &     14.60 & 61.97 & 89.50 \\
            
			\bottomrule
		\end{tabular}
	\caption{
		 Results of the DTI triplet discovery on DrugBank and TTD. 
		 ``CasRel'' and ``Pure NER'' are extractive methods leveraging BERT, and the remaining are generative ones.  
		 ``-attn'' denotes using pre-trained language models in the attention manner.
	}\label{tab:all_results_model}
\end{table*}

\noindent{\bf Hyperparameters}
For CasRel, we mainly follow the hyperparameters suggested by \cite{wei2020novel}.
A modification to CasRel is that since our input text can be longer than $512$ \footnote{After BPE, there are $762$ abstract longer than $512$ tokens, and $19$ abstract longer than $1k$ tokens}, we cut the document into several pieces, each with a length of $512$.
We use BERT to encode each piece and concatenate all the representations for further processing.
We use $66$ relation-classifiers in total, where each classifier is a single-layer feed-forward network with ReLU activation that taking BERT embedding as input. The drug and target identifier is a single-layer feed-forward network.

For the pure NER model,  an entity classifier of the single-layer feed-forward network is applied after the BERT module. We jointly finetune the BERT and the classification heads. We use Adam optimizer \cite{kingma2014adam} with learning rate $10^{-5}$. The minibatch size is $2$ sequences. The models are trained for $100$ epochs with early stopping, where the training stops if the validation performance does not increase for $5$ epochs.


For generative models, after tokenization, we apply BPE~\cite{sennrich2015neural} to both the source sequences and target sequences to reduce vocabularies. 
We set the number of layers as $2$, and the embedding dimension as $256$.
We use Adam optimizer with the \texttt{inverse\_sqrt} scheduler. The learning rate is $5\times10^{-4}$ and warm-up steps are $8k$. The dropout and attention dropout of Transformer are set as $0.2$ and $0.1$. The label smoothing is set as $0.2$. 
The batch size is $12k$ tokens per GPU. 
For the Transformer with pre-trained models, we explore two methods as introduced in \S\ref{sec:end2end}. We try the conventional BERT$_{\rm base}$ model and PubMedBERT$_{\rm base}$ model, in which PubMedBERT$_{\rm base}$ is trained using abstracts of all PubMed papers.
All models are trained on a single V100 GPU.

\noindent{\bf Evaluation metrics}\label{sec:metric}
We define a set of metrics to evaluate the performance of a model for DTI discovery, covering different granularity:  (1) \textit{triplet-level metrics}, which are used in previous knowledge discovery work~\cite{yao2019docred,maimon2005data} and assess the correctness of a discovered triplet as a whole; (2) \textit{ontology-level metrics}, which evaluate the correctness and completeness of all discovered DTI triplets from a paper corpus as a whole; and (3) \textit{entity-level metrics}, which evaluate the accuracy of discovered drugs, targets, and interactions respectively.
We describe the detailed definition of metrics in Appendix C.

\subsection{Results of the End-to-End Method} \label{sec:results_ours}
The test results of DrugBank and TTD are reported in Table~\ref{tab:all_results_model}. Due to space limits, we leave the standard deviation of results in Appendix E and the case study in Appendix F. We have the following observations: 

\noindent(1) Generative methods obtain better results than the extractive method (i.e., CasREL) on \ourDB{}, in terms of triplet-level metric and ontology-level metric. One reason is that our task lacks manual annotation of intermediate labels such as the BIO representations of all entities and relations among any two entities. We obtain such intermediate labels with FuzzyMatch, which are usually of poor quality and therefore impair the performance of extractive methods. For the DTI triplet discovery task, such intermediate labels are often hard to obtain, and we should keep exploring how to improve performances without intermediate labels. 

\noindent(2) For extractive methods, the pure NER method outperforms CasRel on triplet-level metric and entity-level metric. Specifically, for entity-level drug accuracy, the pure NER method even achieves the best result. This shows when intermediate labels are lacking and the relations among entities are comprehensive, simplifying this problem (like extracting only one triplet for a document) is another choice.

\noindent(3) Using pre-trained models is helpful for our task. Taking DrugBank as an example, for triplet-level F1, after using conventional BERT to initialize the encoder, the metric can be improved from $27.41$ to $30.32$. After using PubMedBERT, which is a model pre-trained on all abstracts of PubMed, we achieve an even higher F1 score, $34.82$. This demonstrates the effectiveness of pre-training, especially in-domain pre-training. 

\noindent(4) The manner of using pre-trained models also matters. Comparing with directly initializing the encoder with a pre-trained model, we find that indirectly incorporating the pre-training model can further boost the performance: BERT-attn and PubMedBERT-attn obtain more than $4$ and $2$ point improvement over BERT and PubMedBERT respectively.

\noindent(5) The scores on TTD are lower than DrugBank because TTD is a harder dataset. To verify this, we calculate the minimal distance between drugs and targets: Given a document $D$ and a DTI triplet $(d,t,i)$, let $P_d$ and $P_t$ denote two sets which are positions of drugs and targets obtained by \texttt{FuzzyMatch} in $D$. The distance is defined as $\min_{p_d\in P_d,p_t\in P_t} |p_d-p_t|$. For DrugBank and TTD, the average minimal distances over all test samples are $34$ and $51$, which shows that identifying the DTI triplet from TTD requires understanding a longer document.

\noindent(6) While using pre-trained models achieves the best results, we observe that it suffers from overfitting: The F1 score on the training set is $70.29$ for PubMedBERT, which is much higher than those on the validation set ($23.33$) and test set ($22.9$, the average score of two test sets). We find that simply using larger dropout or label smoothing does not help, which suggests better regularization techniques are needed for this task. More details are in Appendix D.

\paragraph{Effect of generation order.}
Our methods learn to generate drug-target-interaction triplets sequentially for generative methods. An advantage of this method is that we could leverage the dependency among the triplets to improve the generation quality. A question arises: does the order of elements in DTI triplet matter?
To find it out, we enumerate all six orders of the triplet on the standard Transformer model and the 
PubMedBERT-attn model. The results are in Table \ref{tab:order}.
Generally, the order of (drug, interaction, target) performs better, indicating that the order of triplet should be consistent with natural language order (i.e., subject-verb-object).



\begin{table*}[t]
\centering
\footnotesize
\renewcommand\arraystretch{1.2}
\begin{tabular}{ll rrr rrr}
\toprule
\multicolumn{2}{c}{\textbf{Triplet Order}} & D-I-T & D-T-I & I-D-T & I-T-D & T-I-D & T-D-I  \\
\midrule
\multirow{2}{*}{\textbf{DrugBank}}  & Transformer   & \textbf{27.41} & 26.34 & 26.66 & 25.15 & 25.38 & 25.18  \\
            & Transformer + PubMedBERT-attn   & \textbf{36.97} & 36.13 & 32.48 & 33.75 & 34.89 & 36.54 \\
\midrule

\multirow{2}{*}{\textbf{TTD}}   & Transformer   & \textbf{6.32} & 5.01 & 5.87 & 4.72 & 5.81 & 4.37 \\
        & Transformer + PubMedBERT-attn    & \textbf{8.88} & 8.52 & 7.83 & 7.28 & 8.42 & 7.12 \\

\bottomrule
\end{tabular}
\caption{
Results of the generation with different triplet orders.
}\label{tab:order}
\end{table*}

\subsection{Results of the Semi-supervised Method}\label{sec:results_semi}
As generative models perform better than extractive ones, we focus on generative ones in this sub-section and conduct experiments with the Transformer model and Transformer + PubMedBERT-attn model. We merge the KD-DTI corpus with $\mathcal{D}_{\rm DS}$ and $\mathcal{D}_{\rm KD}$ respectively to get two enlarged datasets, and then train models on them. Instead of training from scratch, we find that initializing the parameters from a model trained on the parallel corpus KD-DTI is better. 

The results are shown in Table \ref{tbl:semi}. We have the following observations:

\begin{table*}[t]
\centering
\footnotesize
\renewcommand\arraystretch{1.2}
\begin{tabular}{l ccc ccc}
\toprule
\multirow{2}{*}{\textbf{DrugBank}} & \multicolumn{3}{c}{\textbf{Triplet Level}} & \multicolumn{3}{c}{\textbf{Ontology Level}}  \\
\cmidrule(lr){2-4}
\cmidrule(lr){5-7}
& \multicolumn{1}{c}{\textbf{No Enhance} } & \multicolumn{1}{c}{\textbf{+ DS} } & \multicolumn{1}{c}{\textbf{+ Ours} }  
& \multicolumn{1}{c}{\textbf{No Enhance} } & \multicolumn{1}{c}{\textbf{+ DS} } & \multicolumn{1}{c}{\textbf{+ Ours} } \\
\midrule

Transformer   & 27.41 & 29.92 & 30.57    & 25.49 & 27.26 & 28.04  \\
Transformer + PubMedBERT-attn   & 36.97 & 35.11 & 39.78     & 34.32 & 35.15 & 38.87 \\
 
  \midrule
 \multirow{2}{*}{\textbf{TTD}} & \multicolumn{3}{c}{\textbf{Triplet Level}} & \multicolumn{3}{c}{\textbf{Ontology Level}}  \\

\cmidrule(lr){2-4}
\cmidrule(lr){5-7}
& \multicolumn{1}{c}{\textbf{No Enhance} } & \multicolumn{1}{c}{\textbf{+ DS} } & \multicolumn{1}{c}{\textbf{+ Ours} }  
& \multicolumn{1}{c}{\textbf{No Enhance} } & \multicolumn{1}{c}{\textbf{+ DS} } & \multicolumn{1}{c}{\textbf{+ Ours} } \\
\midrule
Transformer   & 6.32 & 6.99 & 7.20     & 5.75 & 6.19 & 6.66  \\
Transformer + PubMedBERT-attn   & 8.88 & 10.83 & 12.26      & 7.87 & 8.01 & 9.80 \\

\bottomrule
\end{tabular}
\caption{
Comparison of semi-supervised methods.
}\label{tbl:semi}
\end{table*}

\begin{table}[t]
\centering
\footnotesize
\renewcommand\arraystretch{1.2}
\begin{tabular}{l ccc}
\toprule
& \multicolumn{1}{c}{\textbf{DrugBank} } & \multicolumn{1}{c}{\textbf{TTD} } & \multicolumn{1}{c}{\textbf{Average} } \\
\midrule
Our method   & 39.78 & 12.26 & 26.02 \\
~~w/o rule-filter & 33.45 {\scriptsize(-6.33)} & 6.82 {\scriptsize (-5.44)} & 20.14 {\scriptsize(-5.88)} \\
~~w/o model-label & 33.55 {\scriptsize(-6.32)} & 11.09 {\scriptsize(-1.17)} & 22.32 {\scriptsize(-3.70)} \\
DS w/ interaction & 34.04 {\scriptsize(-5.74)} & 12.16 {\scriptsize(-0.10)} & 23.10  {\scriptsize(-2.92)} \\
\bottomrule
\end{tabular}
\caption{
Analysis for the semi-supervised learning methods. We choose the Transformer + PubMedBERT-attn model and report the  triplet F1 scores.
}\label{tbl:semi_ana}
\end{table}

\noindent(1) Enhanced with our semi-supervised method ($\mathcal{D}_{\rm KD}$), we achieve more than 2 points improvement on DrugBank, for both Transformer and Transformer + PubMedBERT; On TTD, significant improvements are also observed. 

\noindent(2) Enhanced with $\mathcal{D}_{\rm DS}$, the generation performance is also generally improved, but not as much as $\mathcal{D}_{\rm KD}$, which shows that the quality of the synthetic data is not as good as that from knowledge distillation. This is consistent with the discovery in~\cite{yao2019docred}. Our conjecture is that the documents are rich in entities and noises, and simply using distance supervision without a scoring mechanism cannot lead to significant improvement, especially when the model equips pre-trained knowledge.

From observations (1) and (2), we can also conclude that pre-training and assigning pseudo labels to the unlabeled data are two orthogonal ways, both of which deserve more attention in the future.



\noindent(3) We also directly combine $\mathcal{D}_{\rm semi}$ with the parallel KD-DTI dataset (which is up sampled by five times) and get the largest training dataset in our experiments. However, while training Transformer (without BERT) with this large dataset, the triplet-level F1 scores on DrugBank and TTD are $18.19$ and $3.15$ respectively, which are much worse than training on KD-DTI only. This demonstrates the necessity of quality control in data enhancement.

\noindent(4) Even if data enhancement can boost DTI discovery, the overall accuracy is still not very high. 
That is, DTI discovery is a challenging task. We need to design better models, algorithms, and/or semi-supervised learning methods to meet the expectation of real-world applications. 

\paragraph{Ablation analysis}
To better understand the proposed semi-supervised method, we remove the \textit{rule-based filtering} step and \textit{model-based labeling} step respectively.
The result is shown in Table \ref{tbl:semi_ana}.
We can see that when removing any of the proposed filtering steps, the performance drops significantly on both DrugBank and TTD data. 
These ablations show the effectiveness of two proposed filtering mechanisms and demonstrate a great necessity for reducing noise in semi-supervised data. We also compare with a variant of distance supervision, where all of drug, target and interaction should appear in the input document (denoted as ``DS w/ interaction'' in Table~\ref{tbl:semi_ana}). We can see such a variant of DS does not bring significant improvement compared to the standard DS.

\section{Related Work}
Early research efforts on knowledge discovery are mainly extractive methods \cite{li2014incremental,miwa2016end,zhong2020frustratingly} and focus on discovering knowledge within a few sentences ~\cite{Zeng2014RelationCV,miwa2016end,Alt2019ImprovingRE,Zhang2019ERNIEEL}.
For document level knowledge discovery, existing solutions are mainly graph-based methods that connect entities across sentences to handle knowledge expressed by multiple sentences \cite{Quirk2017DistantSF,Peng2017CrossSentenceNR,Verga2018SimultaneouslyST,Christopoulou2019ConnectingTD,Nan2020ReasoningWL}.
These methods often require detailed annotations, such as entity mentions and relations between mentions pairs, which are hard to obtain for DTI knowledge discovery.
Recently, generative methods for knowledge discovery are explored \cite{zeng2020copymtl,ye2020contrastive}, which directly generate knowledge triplets from input sentences and only require end-to-end annotations.
However, these works focus on sentence-level extraction. End-to-end generative extraction for document-level text like literature is still less investigated.

For discovering knowledge triplets from literature, early works attempt to extract simple relation between bio-entities. 
\citet{li2016biocreative} extract relation between chemical substances and diseases, and \cite{wu2019renet} focus relation between genes and diseases. 
The genes, diseases, and chemical substances in those work are easier to recognize, and the extracted relationships are relatively simple (only one relation type)
Different from them, DTI covers much more diverse entity terms and more relations.
\citet{chemprot} and \citet{doring2020automated} discover chemical proteins relation on document-level and sentence-level respectively, which are related to DTI knowledge.
However, both of them mainly focus on relation extraction, where the entities are given in advance. By contrast, we aim to jointly discover the DTI triplets from the document. On the other hand, our datasets have more target and relational types and are much larger in volume.

\section{Conclusions and Future Directions}\label{sec:conclusion}
In this work, we explore the first end-to-end generative solution for extracting $\langle$drug, target, interaction$\rangle$ triplets from biomedical literature, which is one of the most important knowledge discovery tasks in the biomedical domain. Besides, we created KD-DTI, the largest dataset for this task, which is expected to boost and advance future research.

There are multiple directions to explore for this task:

\noindent(1) {\em Accuracy improvement}: We have shown that the performance of several state-of-the-art models is still far from industry demand. Therefore,  how to improve accuracy for the task is an important research problem. As shown in this paper, designing better generative models and combing with pre-trained models properly are promising directions. How to effectively leverage unlabeled data (beyond pre-training) is also worthy of exploration. 

\noindent(2) {\em Dataset improvement}: We have created the largest dataset for the DTI discovery task. The dataset can be improved in terms of scale and quality.  Furthermore, there are many other knowledge discovery tasks in the biomedical domain, which also need public datasets for algorithm evaluation.



\bibliography{reference}

\begin{thebibliography}{40}
\providecommand{\natexlab}[1]{#1}

\bibitem[{Alt, H{\"u}bner, and Hennig(2019)}]{Alt2019ImprovingRE}
Alt, C.; H{\"u}bner, M.; and Hennig, L. 2019.
\newblock Improving Relation Extraction by Pre-trained Language
  Representations.
\newblock In \emph{AKBC}.

\bibitem[{Antunes and Matos(2019)}]{chemprot}
Antunes, R.; and Matos, S. 2019.
\newblock Extraction of chemical--protein interactions from the literature
  using neural networks and narrow instance representation.
\newblock \emph{Database}, 2019(baz095).

\bibitem[{Christopoulou, Miwa, and
  Ananiadou(2019)}]{Christopoulou2019ConnectingTD}
Christopoulou, F.; Miwa, M.; and Ananiadou, S. 2019.
\newblock Connecting the Dots: Document-level Neural Relation Extraction with
  Edge-oriented Graphs.
\newblock In \emph{EMNLP-IJCNLP}.

\bibitem[{Devlin et~al.(2019)Devlin, Chang, Lee, and
  Toutanova}]{devlin2018bert}
Devlin, J.; Chang, M.-W.; Lee, K.; and Toutanova, K. 2019.
\newblock Bert: Pre-training of deep bidirectional transformers for language
  understanding.
\newblock \emph{NAAcl}.

\bibitem[{Dong et~al.(2019)Dong, Yang, Wang, Wei, Liu, Wang, Gao, Zhou, and
  Hon}]{UNILM}
Dong, L.; Yang, N.; Wang, W.; Wei, F.; Liu, X.; Wang, Y.; Gao, J.; Zhou, M.;
  and Hon, H. 2019.
\newblock Unified Language Model Pre-training for Natural Language
  Understanding and Generation.
\newblock In Wallach, H.~M.; Larochelle, H.; Beygelzimer, A.;
  d'Alch{\'{e}}{-}Buc, F.; Fox, E.~B.; and Garnett, R., eds., \emph{NeurIPS},
  13042--13054.

\bibitem[{D{\"o}ring et~al.(2020)D{\"o}ring, Qaseem, Becer, Li, Mishra, Gao,
  Kirchner, Sauter, Telukunta, Moumbock et~al.}]{doring2020automated}
D{\"o}ring, K.; Qaseem, A.; Becer, M.; Li, J.; Mishra, P.; Gao, M.; Kirchner,
  P.; Sauter, F.; Telukunta, K.~K.; Moumbock, A.~F.; et~al. 2020.
\newblock Automated recognition of functional compound-protein relationships in
  literature.
\newblock \emph{Plos one}, 15(3): e0220925.

\bibitem[{Gardent et~al.(2017)Gardent, Shimorina, Narayan, and
  Perez-Beltrachini}]{gardent2017creating}
Gardent, C.; Shimorina, A.; Narayan, S.; and Perez-Beltrachini, L. 2017.
\newblock Creating training corpora for nlg micro-planning.
\newblock In \emph{ACL}.

\bibitem[{Gu et~al.(2020)Gu, Tinn, Cheng, Lucas, Usuyama, Liu, Naumann, Gao,
  and Poon}]{pubmedbert}
Gu, Y.; Tinn, R.; Cheng, H.; Lucas, M.; Usuyama, N.; Liu, X.; Naumann, T.; Gao,
  J.; and Poon, H. 2020.
\newblock Domain-Specific Language Model Pretraining for Biomedical Natural
  Language Processing.

\bibitem[{Hinton, Vinyals, and Dean(2015)}]{hinton2015distilling}
Hinton, G.; Vinyals, O.; and Dean, J. 2015.
\newblock Distilling the knowledge in a neural network.
\newblock \emph{arXiv preprint arXiv:1503.02531}.

\bibitem[{Kingma and Ba(2014)}]{kingma2014adam}
Kingma, D.~P.; and Ba, J. 2014.
\newblock Adam: A method for stochastic optimization.
\newblock \emph{arXiv preprint arXiv:1412.6980}.

\bibitem[{Landhuis(2016)}]{Landhuis2016}
Landhuis, E. 2016.
\newblock Scientific literature: Information overload.
\newblock \emph{Nature}, 535(7612): 457--458.

\bibitem[{Li et~al.(2016)Li, Sun, Johnson, Sciaky, Wei, Leaman, Davis,
  Mattingly, Wiegers, and Lu}]{li2016biocreative}
Li, J.; Sun, Y.; Johnson, R.~J.; Sciaky, D.; Wei, C.-H.; Leaman, R.; Davis,
  A.~P.; Mattingly, C.~J.; Wiegers, T.~C.; and Lu, Z. 2016.
\newblock BioCreative V CDR task corpus: a resource for chemical disease
  relation extraction.
\newblock \emph{Database}, 2016.

\bibitem[{Li and Ji(2014)}]{li2014incremental}
Li, Q.; and Ji, H. 2014.
\newblock Incremental joint extraction of entity mentions and relations.
\newblock In \emph{ACL}, 402--412.

\bibitem[{Maimon and Rokach(2005)}]{maimon2005data}
Maimon, O.; and Rokach, L. 2005.
\newblock Data mining and knowledge discovery handbook.

\bibitem[{Mintz et~al.(2009)Mintz, Bills, Snow, and
  Jurafsky}]{mintz2009distant}
Mintz, M.; Bills, S.; Snow, R.; and Jurafsky, D. 2009.
\newblock Distant supervision for relation extraction without labeled data.
\newblock In \emph{ACL}, 1003--1011.

\bibitem[{Miwa and Bansal(2016)}]{miwa2016end}
Miwa, M.; and Bansal, M. 2016.
\newblock End-to-end relation extraction using lstms on sequences and tree
  structures.
\newblock \emph{ACL}.

\bibitem[{Nan et~al.(2020)Nan, Guo, Sekulic, and Lu}]{Nan2020ReasoningWL}
Nan, G.; Guo, Z.; Sekulic, I.; and Lu, W. 2020.
\newblock Reasoning with Latent Structure Refinement for Document-Level
  Relation Extraction.
\newblock In \emph{ACL}.

\bibitem[{Neumann et~al.(2019)Neumann, King, Beltagy, and
  Ammar}]{neumann-etal-2019-scispacy}
Neumann, M.; King, D.; Beltagy, I.; and Ammar, W. 2019.
\newblock {S}cispa{C}y: {F}ast and {R}obust {M}odels for {B}iomedical {N}atural
  {L}anguage {P}rocessing.
\newblock In \emph{Proceedings of the 18th BioNLP Workshop and Shared Task},
  319--327. Florence, Italy: Association for Computational Linguistics.

\bibitem[{Peng et~al.(2017)Peng, Poon, Quirk, Toutanova, and
  Yih}]{Peng2017CrossSentenceNR}
Peng, N.; Poon, H.; Quirk, C.; Toutanova, K.; and Yih, W.-t. 2017.
\newblock Cross-Sentence N-ary Relation Extraction with Graph LSTMs.
\newblock \emph{Transactions of the Association for Computational Linguistics},
  5: 101--115.

\bibitem[{Quirk and Poon(2017)}]{Quirk2017DistantSF}
Quirk, C.; and Poon, H. 2017.
\newblock Distant Supervision for Relation Extraction beyond the Sentence
  Boundary.
\newblock In \emph{EACL}.

\bibitem[{Radford et~al.(2019)Radford, Wu, Child, Luan, Amodei, and
  Sutskever}]{radford2019language}
Radford, A.; Wu, J.; Child, R.; Luan, D.; Amodei, D.; and Sutskever, I. 2019.
\newblock Language models are unsupervised multitask learners.
\newblock \emph{OpenAI Blog}, 1(8): 9.

\bibitem[{Riedel, Yao, and McCallum(2010)}]{riedel2010modeling}
Riedel, S.; Yao, L.; and McCallum, A. 2010.
\newblock Modeling relations and their mentions without labeled text.
\newblock In \emph{Joint European Conference on Machine Learning and Knowledge
  Discovery in Databases}, 148--163. Springer.

\bibitem[{Sachdev and Gupta(2019)}]{SACHDEV2019103159}
Sachdev, K.; and Gupta, M.~K. 2019.
\newblock A comprehensive review of feature based methods for drug target
  interaction prediction.
\newblock \emph{Journal of Biomedical Informatics}, 93: 103159.

\bibitem[{Sennrich, Haddow, and Birch(2015)}]{sennrich2015neural}
Sennrich, R.; Haddow, B.; and Birch, A. 2015.
\newblock Neural machine translation of rare words with subword units.
\newblock \emph{arXiv preprint arXiv:1508.07909}.

\bibitem[{Vaswani et~al.(2017)Vaswani, Shazeer, Parmar, Uszkoreit, Jones,
  Gomez, Kaiser, and Polosukhin}]{Transformer}
Vaswani, A.; Shazeer, N.; Parmar, N.; Uszkoreit, J.; Jones, L.; Gomez, A.~N.;
  Kaiser, {\L}.; and Polosukhin, I. 2017.
\newblock Attention is all you need.
\newblock In \emph{NIPS}, 5998--6008.

\bibitem[{Verga, Strubell, and McCallum(2018)}]{Verga2018SimultaneouslyST}
Verga, P.; Strubell, E.; and McCallum, A. 2018.
\newblock Simultaneously Self-Attending to All Mentions for Full-Abstract
  Biological Relation Extraction.
\newblock In \emph{NAACL-HLT}.

\bibitem[{Wang et~al.(2020)Wang, Zhang, Li, Zhou, Zhang, Wang, Zhang, Zhu, Ren,
  Tan et~al.}]{wang2020therapeutic}
Wang, Y.; Zhang, S.; Li, F.; Zhou, Y.; Zhang, Y.; Wang, Z.; Zhang, R.; Zhu, J.;
  Ren, Y.; Tan, Y.; et~al. 2020.
\newblock Therapeutic target database 2020: enriched resource for facilitating
  research and early development of targeted therapeutics.
\newblock \emph{Nucleic acids research}, 48(D1): D1031--D1041.

\bibitem[{Wei et~al.(2020)Wei, Su, Wang, Tian, and Chang}]{wei2020novel}
Wei, Z.; Su, J.; Wang, Y.; Tian, Y.; and Chang, Y. 2020.
\newblock A novel cascade binary tagging framework for relational triple
  extraction.
\newblock In \emph{ACL}, 1476--1488.

\bibitem[{Wishart et~al.(2017)Wishart, Feunang, Guo, Lo, Marcu, Grant, Sajed,
  Johnson, Li, Sayeeda, Assempour, Iynkkaran, Liu, Maciejewski, Gale, Wilson,
  Chin, Cummings, Le, Pon, Knox, and Wilson}]{Wishart2017}
Wishart, D.~S.; Feunang, Y.~D.; Guo, A.~C.; Lo, E.~J.; Marcu, A.; Grant, J.~R.;
  Sajed, T.; Johnson, D.; Li, C.; Sayeeda, Z.; Assempour, N.; Iynkkaran, I.;
  Liu, Y.; Maciejewski, A.; Gale, N.; Wilson, A.; Chin, L.; Cummings, R.; Le,
  D.; Pon, A.; Knox, C.; and Wilson, M. 2017.
\newblock {DrugBank} 5.0: a major update to the {DrugBank} database for 2018.
\newblock \emph{Nucleic Acids Research}, 46(D1): D1074--D1082.

\bibitem[{Wu et~al.(2019)Wu, Luo, Leung, Ting, and Lam}]{wu2019renet}
Wu, Y.; Luo, R.; Leung, H.~C.; Ting, H.-F.; and Lam, T.-W. 2019.
\newblock Renet: A deep learning approach for extracting gene-disease
  associations from literature.
\newblock In \emph{International Conference on Research in Computational
  Molecular Biology}, 272--284. Springer.

\bibitem[{Wu et~al.(2018)Wu, Li, Liu, and Tang}]{wu2018network}
Wu, Z.; Li, W.; Liu, G.; and Tang, Y. 2018.
\newblock Network-based methods for prediction of drug-target interactions.
\newblock \emph{Frontiers in pharmacology}, 9: 1134.

\bibitem[{Yamanishi et~al.(2008)Yamanishi, Araki, Gutteridge, Honda, and
  Kanehisa}]{yamanishi2008prediction}
Yamanishi, Y.; Araki, M.; Gutteridge, A.; Honda, W.; and Kanehisa, M. 2008.
\newblock Prediction of drug--target interaction networks from the integration
  of chemical and genomic spaces.
\newblock \emph{Bioinformatics}, 24(13): i232--i240.

\bibitem[{Yao et~al.(2019)Yao, Ye, Li, Han, Lin, Liu, Liu, Huang, Zhou, and
  Sun}]{yao2019docred}
Yao, Y.; Ye, D.; Li, P.; Han, X.; Lin, Y.; Liu, Z.; Liu, Z.; Huang, L.; Zhou,
  J.; and Sun, M. 2019.
\newblock DocRED: A large-scale document-level relation extraction dataset.
\newblock \emph{ACL}.

\bibitem[{Ye et~al.(2021)Ye, Zhang, Deng, Chen, Tan, Huang, and
  Chen}]{ye2020contrastive}
Ye, H.; Zhang, N.; Deng, S.; Chen, M.; Tan, C.; Huang, F.; and Chen, H. 2021.
\newblock Contrastive Triple Extraction with Generative Transformer.
\newblock In \emph{AAAI}.

\bibitem[{Zeng et~al.(2014)Zeng, Liu, Lai, Zhou, and Zhao}]{Zeng2014RelationCV}
Zeng, D.; Liu, K.; Lai, S.; Zhou, G.; and Zhao, J. 2014.
\newblock Relation Classification via Convolutional Deep Neural Network.
\newblock In \emph{COLING}.

\bibitem[{Zeng, Zhang, and Liu(2020)}]{zeng2020copymtl}
Zeng, D.; Zhang, R.~H.; and Liu, Q. 2020.
\newblock CopyMTL: Copy Mechanism for Joint Extraction of Entities and
  Relations with Multi-Task Learning.
\newblock In \emph{AAAI}.

\bibitem[{Zhang and Zhou(2013)}]{zhang2013review}
Zhang, M.-L.; and Zhou, Z.-H. 2013.
\newblock A review on multi-label learning algorithms.
\newblock \emph{IEEE transactions on knowledge and data engineering}, 26(8):
  1819--1837.

\bibitem[{Zhang et~al.(2019)Zhang, Han, Liu, Jiang, Sun, and
  Liu}]{Zhang2019ERNIEEL}
Zhang, Z.; Han, X.; Liu, Z.; Jiang, X.; Sun, M.; and Liu, Q. 2019.
\newblock ERNIE: Enhanced Language Representation with Informative Entities.
\newblock In \emph{ACL}.

\bibitem[{Zhong and Chen(2020)}]{zhong2020frustratingly}
Zhong, Z.; and Chen, D. 2020.
\newblock A Frustratingly Easy Approach for Joint Entity and Relation
  Extraction.
\newblock \emph{arXiv preprint arXiv:2010.12812}.

\bibitem[{Zhu et~al.(2020)Zhu, Xia, Wu, He, Qin, Zhou, Li, and
  Liu}]{DBLP:conf/iclr/ZhuXWHQZLL20}
Zhu, J.; Xia, Y.; Wu, L.; He, D.; Qin, T.; Zhou, W.; Li, H.; and Liu, T. 2020.
\newblock Incorporating {BERT} into Neural Machine Translation.
\newblock In \emph{ICLR}.

\end{thebibliography}

\clearpage

\appendix

\section*{Appendix}

\section{Dataset Introduction}
\noindent(1) Our dataset, \ourDB{}, is about to speed up the research of discovering drug, target and their interaction from the literature, which is an important topic. 
Our dataset is built upon DrugBank and TTD. After the confirmation from the owners of DrugBank, to use our dataset, one should register an account for DrugBank to extract the drug and target names related to DrugBank. 
For TTD, we confirm with the owner, and no license is required.

The dataset is in JSON format, which can be directly loaded. The data structure of our dataset is shown as follows: 
\begin{figure}[!h]
	\centering
	\begin{tikzpicture}
	\draw (0,0 ) node[inner sep=0] {\includegraphics[width=1\columnwidth, trim={0cm 6cm 14.5cm 0cm}, clip]{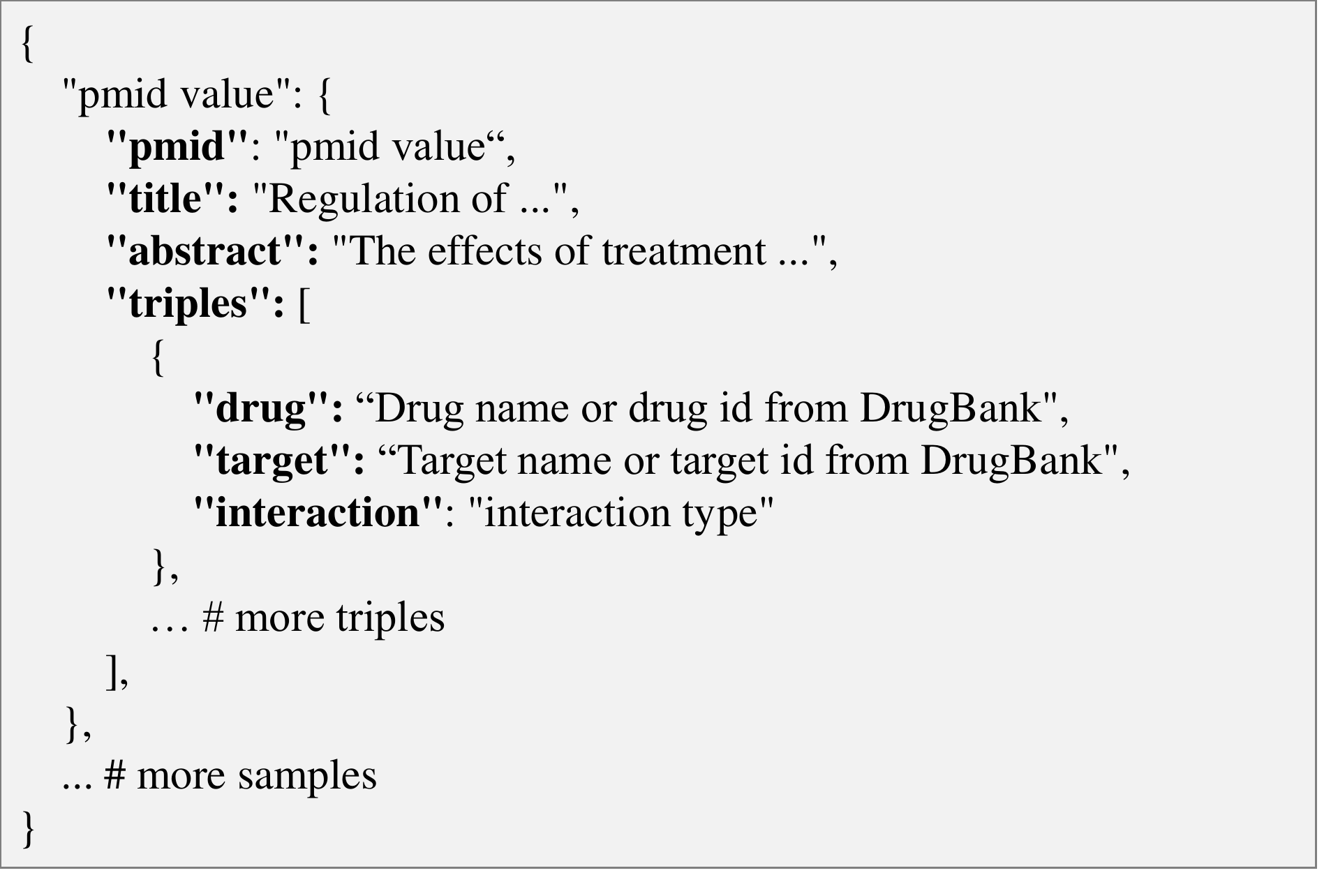}};
	\end{tikzpicture}
\end{figure}

\noindent(2) Any researchers about machine learning, natural language processing, biology and medicine might benefit from our dataset.

\noindent(3) Currently, the dataset is only visible to reviewers through the private URL.  After the review process, we will release our dataset through Github or  a publicly available website. Our dataset will be maintained for a long time.

\noindent(4) We have confirmed with the owners of DrugBank and TTD for re-distribution.

\noindent(5) The licence of the dataset is the Computational Use of Data Agreement (C-UDA) License.

\noindent(6) We will update our dataset regularly according to the feedback of users.

\section{Detailed Process for Data Filtration.}
For ease of reference, we denote the dataset obtained without filtration as $\mathcal{D} = \{D_j, \{Y_{j,k}\}_{k=1}^{K_j}\}$,  where (1) $D_j$ is the document (i.e., title and abstract); (2) $Y_{j,k}=(d_{j,k},t_{j,k},i_{j,k})$ is the $k$-th triplet of $D_j$, with each element representing drug, target and interaction respectively; (3) $K_j$ is the number of triplets associated with $D_j$.

As a starting point of structured DTI knowledge discovery, we are only interested in the document which contains enough information to discover a DTI triplet. However, in $\mathcal{D}$, 
some papers only generally describe some drugs and targets, in which the DTI triplets do not explicitly appear. 
Therefore, we heuristically filter out the samples in $\mathcal{D}$ by which we cannot obtain the associated DTI triplets. 
The basic idea is that we require that the drug, target, and interaction in a triplet should be all included in a paper. We describe the details of the filtration process as below. 



Given a query $q$ and a document $D$, we first use \texttt{FuzzyMatch}\footnote{
An open-sourced tool that leverages Levenshtein distance to fetch similar words to the query. Simple variants are allowed like ``+s'', ``+ed'', etc. \url{https://github.com/taleinat/fuzzysearch}} to retrieve all similar words of $q$ and its synonyms in $D$, and denote them as $\mathcal{R}(q, D)=\{r_j\}_{j=1}^{|\mathcal{R}|}$, where $r_j$ is a retrieved phrase. Here the query can be a drug, a target, or an interaction, and both $q$ and $r_j$ could be a single word or a phrase with multiple words. Note that we obtain synonyms of a drug or target from the Drugbank and TTD database, where entities are attached with synonyms. Based on the retrieval results, we categorize $D$ as one of the following patterns for $q$:
\begin{enumerate}
\item {\em Reliable pattern}, where the query and the fetched words are almost the same;

\item {\em Positive pattern}, where the query and the fetched words share lots of parts in common;

\item {\em Negative pattern}, where the query is not related to the document.
\end{enumerate}



The detailed pattern are summarized in Table~\ref{tab:rules}.

\begin{table*}[!htbp]
\renewcommand{\arraystretch}{1.3}
\centering
\small
\begin{tabular}{ll@{}}
\toprule
\multicolumn{2}{l}{{\em Input:} A query $q$ and the retrival results $\mathcal{R}(q, D)=\{r_j\}$.}\\
\midrule
\multicolumn{2}{l}{{\em Reliable patterns }}\\
P1& $\exists r_j\in\mathcal{R}(q,D)$ s.t. $r_j$ and $q$ are exactly the same \\
&(except for parentheses, cases, and punctuation marks);\\
P2& $\exists r_j\in\mathcal{R}(q,D)$ s.t. $r_j$ and $q$ have $>20$ characters 
or $\ge3$ words in common;\\
\midrule
\multicolumn{2}{l}{{\em Positive patterns }}\\
P3 & There are at least $2$ elements in $\mathcal{R}$ matching variant formats of $q$ (e.g., +s, +ed, +ing);\\
P4 & There are at least $2$ elements $r_\cdot$ in $\mathcal{R}$ s.t each $r$ and $q$ has $>8$ characters in common,  \\ 
& or at most $10\%$ different characters;\\
P5 & $|\mathcal{R}(q,D)|>3$;\\
\midrule
\multicolumn{2}{l}{{\em Negative patterns }}\\
P6 & Each $r$ in $\mathcal{R}$ has $<8$ characters, or $r$  is a meaningless word like ``other'', ``unknown'', etc.\\
\bottomrule
\end{tabular}
\caption{Matching patterns between a query and a paper.}
\label{tab:rules}
\end{table*}

Denote the matching score between a query $q$ and a document $D$ as $\varphi(q,D)$: If $q$ is a reliable pattern of $D$, we set $\varphi(q,D)=5$; if $q$ is a positive pattern, we set $\varphi(q,D)=1$; otherwise, $\varphi(q,D)=-1$. 
We tried several different score setting and determined the scores with best data quality through manual check.
Given any document $D_j$ and its $k$-th DTI triplet $Y_{j,k}=(d_{j,k}, t_{j,k}, i_{j,k})$, the matching score between the triplet and document is calculated as follows: 
\begin{equation*}
c(D_j,Y_{j,k}) = \varphi(d_{j,k},D_j) + \varphi(t_{j,k},D_j) + \varphi(i_{j,k},D_j).
\end{equation*}

We sort all samples according to the  matching scores in descending order, filter out the low-confidence samples whose score is less than zero, 
and only keep the top $14k$ high-confidence documents-triplets pairs.
We pick $1.3k$ documents as the initial test set, the $1k$ as the validation set, and the remaining documents ($12k$) as the training set.

\section{Definition of Evaluation Metrics}
In this section, we present the detailed definition of evaluation metrics for DTI discovery, which cover different granularity:
(1) triplet-level metrics, 
(2) ontology-level metrics, and 
(3) entity-level metrics. 

Let $N$ denote the number of documents/papers in a test set. For the $j$-th test sample/document, the set of its associated DTI triplets is denoted as $Y_j^*$. Let $\hat{Y}_j$ denote the output of a model for the $j$-the sample, which is another set of DTI triplets.

\noindent{\bf Triplet-level metrics}
Following previous work on knowledge extraction~\cite{yao2019docred}, we evaluate that given a document, whether the model could correctly discover the corresponding DTI triplets. Since a single paper may contain multiple DTI triplets, we define precision (P), recall (R) and F1 score~\cite{zhang2013review} as follows:
\begin{flalign*} 
& {\rm P} = \frac{1}{N} \sum_{j=1}^N \frac{|Y^*_j \cap \hat{Y}_j|}{|\hat{Y}_j|},  \;
{\rm R}  = \frac{1}{N} \sum_{j=1}^N \frac{|Y^*_j \cap \hat{Y}_j|}{|Y^*_j|}, \;
{\rm F1} = \frac{2{\rm P}{\rm R} }{{\rm P}+{\rm R} }. 
\end{flalign*} 

\noindent{\bf Ontology-level metrics}
For industrial applications, given a corpus of documents, one of the important objectives is to find out all possible knowledge from the corpus. To evaluate the knowledge coverage from the corpus level, we define ontology-level metrics (i.e., corpus-level metric) that evaluates how many triplets of the entire corpus are correctly extracted.

Define $Y^*=\cup_{j=1}^{N}Y^*_j$ and $\hat{Y}=\cup_{j=1}^{N}\hat{Y}_j$. The ontology level precision (P), recall (R) and F1 are:
\begin{flalign*} 
& {\rm P}  =  \frac{|Y^* \cap \hat{Y}|}{|\hat{Y}|},\,
 {\rm R}  =  \frac{|Y^* \cap \hat{Y}|}{|Y^*|},  \,
{\rm F1} = \frac{2{\rm P} {\rm R}  }{ {\rm P} +{\rm R} }.
\end{flalign*} 

\noindent{\bf Entity-level metrics}
As mentioned before, a biomedical paper often contains lots of entities, but many of them are not related to the DTI triplets we want to discover. It is important to extract the right drugs, targets, and interactions from literature. Therefore, we assess the accuracy for drugs (A$_{\rm d}$), targets (A$_{\rm t}$), interactions (A$_{\rm i}$) respectively. 

Let $D^*_j$ and $\hat{D}_j$ denote the sets of all drugs in the ground-truth triplets and model outputs for the $j$-th sample, and similarly for $T^*_j$, $\hat{T}_j$, $I^*_j$, $\hat{I}_j$. We define drug accuracy, target accuracy, and interaction accuracy as below:
\[ 
	\begin{array}{c}
\displaystyle {\rm A}_{\rm d} = \frac{1}{N} \sum_{j=1}^N \frac{|D^*_j \cap \hat{D}_j|}{|D^*_j \cup \hat{D}_j|},  \nonumber 
\displaystyle {\rm A}_{\rm t} = \frac{1}{N} \sum_{j=1}^N \frac{|T_j^* \cap \hat{T}_j|}{|T_j^* \cup \hat{T}_j|}, \nonumber  \\
\displaystyle {\rm A}_{\rm i} = \frac{1}{N} \sum_{j=1}^N \frac{|I_j^* \cap \hat{I}_j|}{|I_j^* \cup \hat{I}_j|}. \nonumber 
	\end{array}
\]






\section{Study on Regularization Techniques}

\begin{table}[!htbp]
\centering
\begin{tabular}{cccccc}
\toprule
(\texttt{dr},\texttt{ls}) & $0.1$ & $0.2$ & $0.3$ \\ 
\midrule
$0.2$ & $18.05$ & $22.24$ & $21.97$ \\
$0.3$ & $20.06$ & $23.33$ & $21.45$ \\
$0.4$ & $20.93$ & $20.14$ & $19.86$ \\
$0.5$ & $18.71$ & $18.97$ & $18.35$ \\
\bottomrule
\end{tabular}
\caption{Results of the F1 on the validation set. \texttt{dr} and \texttt{ls} stand for dropout and labeling smoothing respectively.}
\label{tab:reg_study}
\end{table}

We explore various combinations of dropout and label smoothing based on Transformer + PubMedBERT-attn. The F1 scores of the validation set are reported in Table~\ref{tab:reg_study}. The best result is achieved when dropout and label smoothing are set as $0.3$ and $0.2$ respectively. However, the training F1 score is $70.29$, which is significantly larger than the validation set. We keep increasing dropout and label smoothing, and found that the validation performance cannot be further improved. This shows that using basic techniques to improve generalization (i.e. dropout, label smoothing) can bring limited improvement, and we need more effective regularization techniques for the DTI triplet extraction task.

\section{Main Results with Standard Deviation}
\begin{figure*}[!h]
	\centering
	\begin{tikzpicture}
	\draw (0,0 ) node[inner sep=0] {\includegraphics[width=1.4\columnwidth, trim={0cm 0cm 0cm 0cm}, clip]{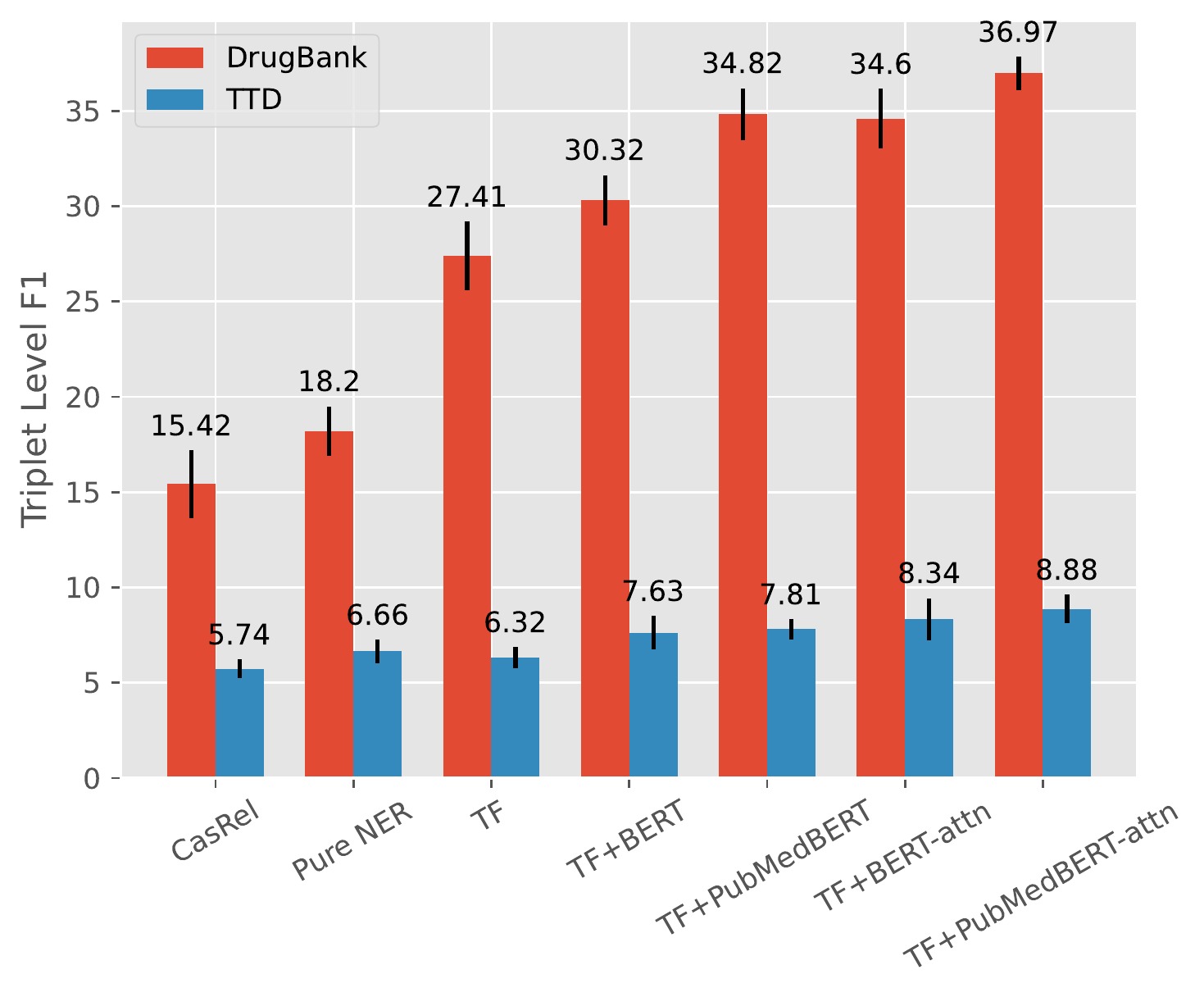}};
	\end{tikzpicture}
	\caption{\footnotesize
		Main results with standard deviation. ``TF'' denotes Transformer.
	 }\label{fig:std}
\end{figure*}

Figure \ref{fig:std} presents the standard deviation of results on DrugBank and TTD. 
On DrugBank, the standard deviation of each model is around $1.5$.
On TTD, the standard deviation scores are smaller and usually less than $1.0$.

\section{Case Study}
We perform a case study to investigate whether a model trained on the KD-DTI dataset is able to discover unseen Drug-Target-Interaction  triplets and handle unseen paper.
To achieve this, we train a generative model on KD-DTI and make predictions on unseen samples from another dataset, ChemProt, which contains human annotation of chemical-protein relation (a kind of drug-target interaction).We use Transformer+PubMedBERT-attn for case study.

As shown in Table \ref{tbl:case}, in the first case, the entire triplet is correctly extracted, while both the drug and whole triplet are unseen in KD-DTI.
In the second case, we successfully extract one of the two annotated triplets. We attribute the missing of the second triplet to the irregular format of drug ``C'' and the presence of distracting items, such as ``P'' and ``M''.

\begin{table*}[t]
\centering
\footnotesize
\renewcommand\arraystretch{1.2}
	\resizebox{\linewidth}{!}{
\begin{tabular}{p{2\columnwidth}}
\toprule
\textbf{Title:} Assessment of the abuse liability of ABT-288, a novel histamine H3 receptor antagonist.
\\
\textbf{Abstract:} RATIONALE: Histamine H3 receptor antagonists, such as ABT-288, have been shown to possess cognitive-enhancing and wakefulness-promoting effects. On the surface, this might suggest that H3 antagonists possess psychomotor stimulant-like effects and, as such, may have the potential for abuse. OBJECTIVES: The aim of the present study was to further characterize whether ABT-288 possesses stimulant-like properties and whether its pharmacology gives rise to abuse liability. METHODS: The locomotor-stimulant effects of ABT-288 were measured in mice and rats, and potential development of sensitization was addressed. Drug discrimination was used to assess amphetamine-like stimulus properties, and drug self-administration was used to evaluate reinforcing effects of ABT-288. The potential development of physical dependence was also studied. RESULTS: ABT-288 lacked locomotor-stimulant effects in both rats and mice. Repeated administration of ABT-288 did not result in cross-sensitization to the stimulant effects of d-amphetamine in mice, suggesting that there is little overlap in circuitries upon which the two drugs interact for motor activity. ABT-288 did not produce amphetamine-like discriminative stimulus effects in drug discrimination studies nor was it self-administered by rats trained to self-administer cocaine. There were no signs of physical dependence upon termination of repeated administration of ABT-288 for 30 days. CONCLUSIONS: The sum of these preclinical data, the first of their kind applied to H3 antagonists, indicates that ABT-288 is unlikely to possess a high potential for abuse in the human population and suggests that H3 antagonists, as a class, are similar in this regard.
\\
\midrule
\textbf{Prediction:~}  (Drug: ``ABT-288'',   Target: ``Histamine H3 receptor'', Interaction: ``antagonist'') \\
\textbf{Annotation:} (Drug: ``ABT-288'',   Target: ``Histamine H3 receptor'', Interaction: ``antagonist'')
\\
\toprule
\textbf{Title:} Mechanisms of Glucose Lowering of Dipeptidyl Peptidase-4 Inhibitor Sitagliptin When Used Alone or With Metformin in Type 2 Diabetes: A double-tracer study.
\\
\textbf{Abstract:}  OBJECTIVE To assess glucose-lowering mechanisms of sitagliptin (S), metformin (M), and the two combined (M+S).RESEARCH DESIGN AND METHODS We randomized 16 patients with type 2 diabetes mellitus (T2DM) to four 6-week treatments with placebo (P), M, S, and M+S. After each period, subjects received a 6-h meal tolerance test (MTT) with [(14)C]glucose to calculate glucose kinetics. Fasting plasma glucose (FPG), fasting plasma insulin, C-peptide (insulin secretory rate [ISR]), fasting plasma glucagon, and bioactive glucagon-like peptide (GLP-1) and gastrointestinal insulinotropic peptide (GIP) was measured.RESULTS FPG decreased from P, 160 $\pm$ 4 to M, 150 $\pm$ 4; S, 154 $\pm$ 4; and M+S, 125 $\pm$ 3 mg/dL. Mean post-MTT PG decreased from P, 207 $\pm$ 5 to M, 191 $\pm$ 4; S, 195 $\pm$ 4; and M+S, 161 $\pm$ 3 mg/dL (P < 0.01]. The increase in mean post-MTT plasma insulin and in ISR was similar in P, M, and S and slightly greater in M+S. Fasting plasma glucagon was equal (65-75 pg/mL) with all treatments, but there was a significant drop during the initial 120 min with S 24\% and M+S 34\% (both P < 0.05) vs. P 17\% and M 16\%. Fasting and mean post-MTT plasma bioactive GLP-1 were higher (P < 0.01) after S and M+S vs. M and P. Basal endogenous glucose production (EGP) fell from P 2.0 $\pm$ 0.1 to S 1.8 $\pm$ 0.1 mg/kg  min, M 1.8 $\pm$ 0.2 mg/kg min [both P < 0.05 vs. P), and M+S 1.5 $\pm$ 0.1 mg/kg  min (P < 0.01 vs. P). Although the EGP slope of decline was faster in M and M+S vs. S, all had comparable greater post-MTT EGP inhibition vs. P (P < 0.05).CONCLUSION SM+S combined produce additive effects to 1) reduce FPG and postmeal PG, 2) augment GLP-1 secretion and $\beta$-cell function, 3) decrease plasma glucagon, and 4) inhibit fasting and postmeal EGP compared with M or S monotherapy.
\\
\midrule
\textbf{Prediction:~}  (Drug: ``Sitagliptin'',   Target: ``Dipeptidyl Peptidase 4'', Interaction: ``inhibitor'') \\
\textbf{Annotation:} (Drug: ``Sitagliptin'',   Target: ``Dipeptidyl Peptidase 4'', Interaction: ``inhibitor''), (Drug: ``C'',   Target: ``C-peptide'', Interaction: ``part of'')
\\
\bottomrule
\end{tabular}
	}
\caption{
Case study of triplet generation on unseen sample.
}\label{tbl:case}
\end{table*}

\section{Broader Impact}
We propose a new dataset for biomedical knowledge discovery. We believe that this dataset can speed up the research of bioNLP and machine learning. For negative impact, after the success of automatic knowledge discovery, it might cause some unemployment of the related researchers and engineers. 

\section{Distribution of Interactions}
\begin{figure*}[!h]
	\centering
	\begin{tikzpicture}
	\draw (0,0 ) node[inner sep=0] {\includegraphics[width=1.6\columnwidth, trim={10cm 1.5cm 10cm 1.5cm}, clip]{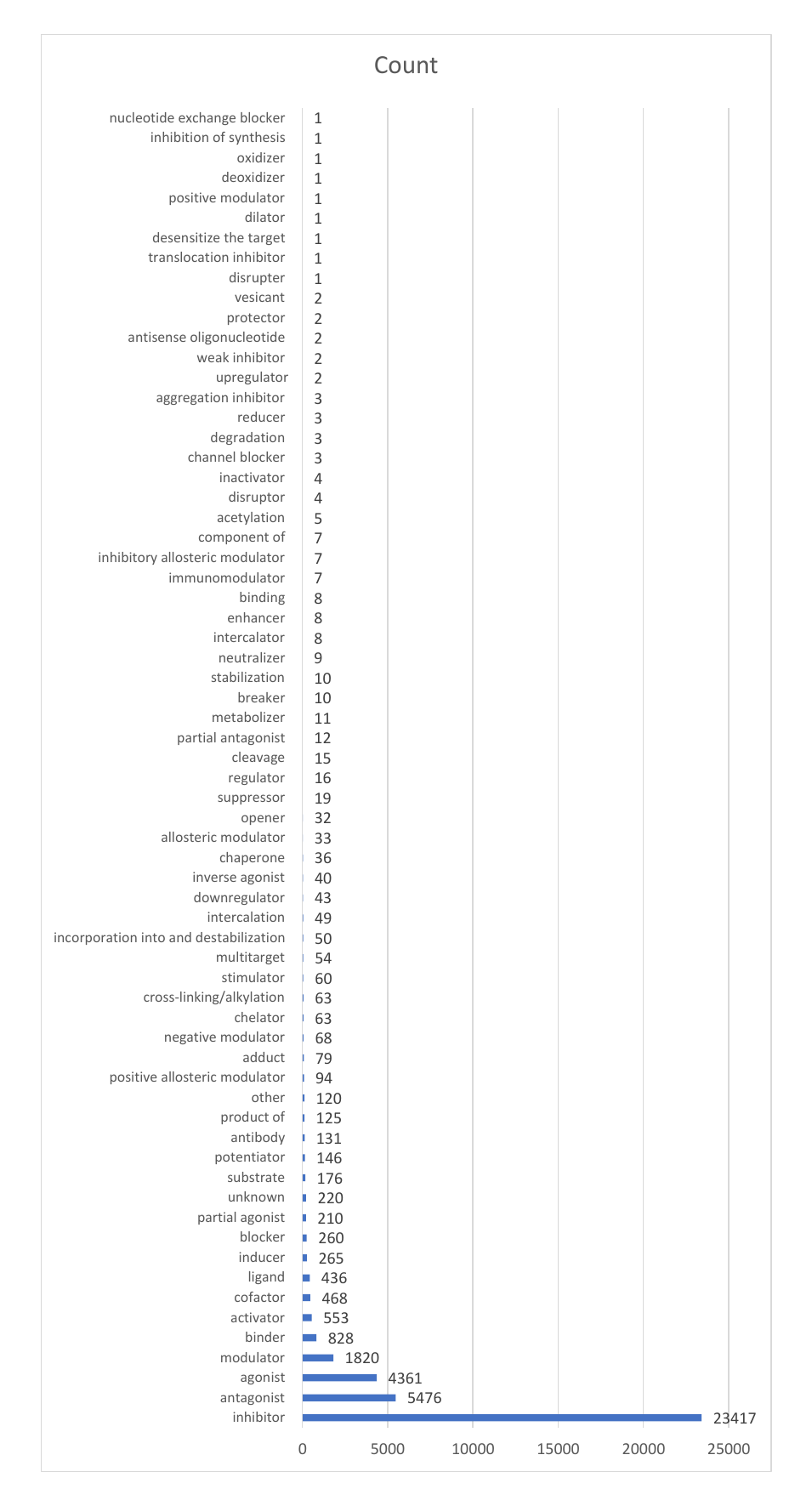}};
	\end{tikzpicture}
	\caption{\footnotesize
		Distribution of interactions.
	 }\label{fig:interaction_s}
\end{figure*}

In the Figure \ref{fig:interaction_s},  we present a details statistic of interactions included in the corpus.

\end{document}